\documentclass[11pt,a4paper]{article}
\usepackage[hyperref]{acl2020}
\usepackage{times}
\usepackage{latexsym}

\usepackage{microtype}

\usepackage{graphicx}
\usepackage{amsmath}
\usepackage{amssymb}
\usepackage{multirow}
\usepackage{array}
\usepackage{algorithm}
\usepackage{algorithmic}
\usepackage{enumitem}

\aclfinalcopy 
\def\aclpaperid{1238} 

\title{Exclusive Hierarchical Decoding for Deep Keyphrase Generation}

\author{Wang Chen\textsuperscript{\rm 1}, 
Hou Pong Chan\textsuperscript{\rm 1}, 
Piji Li\textsuperscript{\rm 2},
Irwin King\textsuperscript{\rm 1} \\
\textsuperscript{\rm 1}The Chinese University of Hong Kong, Shatin, N.T., Hong Kong \\
\textsuperscript{\rm 2}Tencent AI Lab\\
{\tt \textsuperscript{\rm 1}\{wchen, hpchan, king\}@cse.cuhk.edu.hk}\\
{\tt \textsuperscript{\rm 2}pijili@tencent.com}
}

\date{}

\begin{document}
\maketitle
\begin{abstract}
  Keyphrase generation (KG) aims to summarize the main ideas of a document into a set of keyphrases. A new setting is recently introduced into this problem, in which, given a document, the model needs to predict a set of keyphrases and simultaneously determine the appropriate number of keyphrases to produce. Previous work in this setting employs a sequential decoding process to generate keyphrases. However, such a decoding method ignores the intrinsic hierarchical compositionality existing in the keyphrase set of a document. Moreover, previous work tends to generate duplicated keyphrases, which wastes time and computing resources. To overcome these limitations, we propose an exclusive hierarchical decoding framework that includes a hierarchical decoding process and either a soft or a hard exclusion mechanism. The hierarchical decoding process is to explicitly model the hierarchical compositionality of a keyphrase set. Both the soft and the hard exclusion mechanisms keep track of previously-predicted keyphrases within a window size to enhance the diversity of the generated keyphrases. Extensive experiments on multiple KG benchmark datasets demonstrate the effectiveness of our method to generate less duplicated and more accurate keyphrases\footnote{Our code is available at \url{https://github.com/Chen-Wang-CUHK/ExHiRD-DKG}.}.
\end{abstract}

\section{Introduction}
\begin{figure}[t]
\centering
\includegraphics[width=\columnwidth]{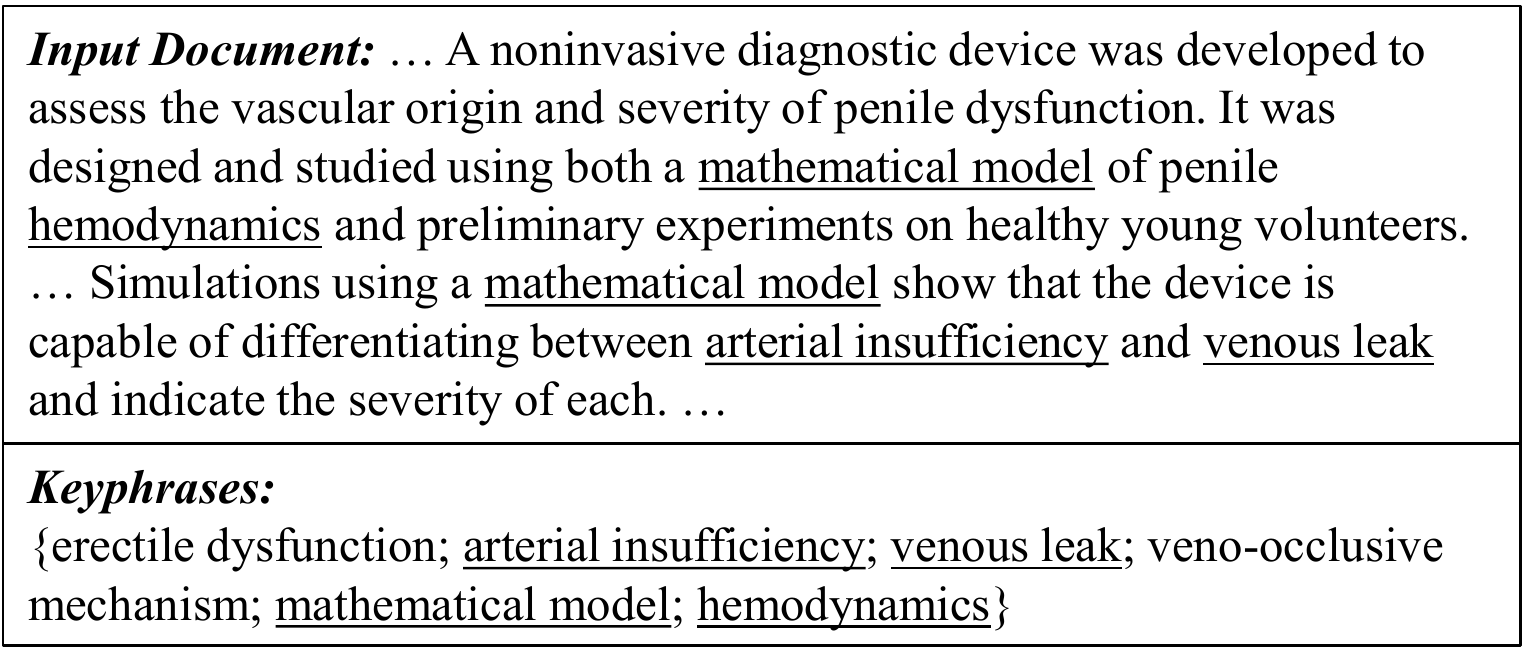}
\caption{An example of an input document and its expected keyphrase output for keyphrase generation problem. Present keyphrases that appear in the document are underlined.}
\label{figure: case_study_intro}
\end{figure}

Keyphrases are short phrases that indicate the core information of a document. As shown in Figure~\ref{figure: case_study_intro}, the keyphrase generation (KG) problem focuses on automatically producing a \emph{keyphrase set} (a set of keyphrases) for the given document. Because of the condensed expression, keyphrases can benefit various downstream applications including opinion mining~\cite{berend2011opinion_opin_app,wilson2005recognizing_senti_app}, document clustering~\cite{hulth2006study_doc_clus_app}, and text summarization~\cite{wang2013domain_summarize_app}.

Keyphrases of a document can be categorized into two groups: \textit{present keyphrase} that appears in the document and \textit{absent keyphrase} that does not appear in the document. Recent generative methods for KG apply the attentional encoder-decoder framework~\cite{luong2015effective,DBLP:conf/iclr/BahdanauCB14} with copy mechanism~\cite{Gu2016copy,see2017get} to predict both present and absent keyphrases. To generate multiple keyphrases for an input document, these methods first use beam search to generate a huge number of keyphrases (e.g., 200) and then pick the top $N$ ranked keyphrases as the final prediction. Thus, in other words, these methods can only predict a fixed number of keyphrases for all documents. 

However, in a practical situation, the appropriate number of keyphrases varies according to the content of the input document. To simultaneously predict keyphrases and determine the suitable number of keyphrases,  \citet{yuan2018generating_diverse} adopts a sequential decoding method with greedy search to generate one sequence consisting of the predicted keyphrases and separators. For example, the produced sequence may be ``hemodynamics [sep] erectile dysfunction [sep] ...'', where ``[sep]'' is the separator. After producing an ending token, the decoding process terminates. The final keyphrase predictions are obtained after splitting the sequence by separators. 
However, there are two drawbacks to this method. First, the sequential decoding method ignores the hierarchical compositionality existing in a keyphrase set (a keyphrase set is composed of multiple keyphrases and each keyphrase consists of multiple words). In this work, we examine the hypothesis that a generative model can predict more accurate keyphrases by incorporating the knowledge of the hierarchical compositionality in the decoder architecture. 
Second, the sequential decoding method tends to generate duplicated keyphrases. It is simple to design specific post-processing rules to remove the repeated keyphrases, but generating and then removing repeated keyphrases wastes time and computing resources. To address these two limitations, we propose a novel exclusive hierarchical decoding framework for KG, which includes a hierarchical decoding process and an exclusion mechanism. 

Our hierarchical decoding process is designed to explicitly model the hierarchical compositionality of a keyphrase set. It is composed of phrase-level decoding (PD) and word-level decoding (WD). A PD step determines which aspect of the document to summarize based on both the document content and the aspects summarized by previously-generated keyphrases. The hidden representation of the captured aspect is employed to initialize the WD process. Then, a new WD process is conducted under the PD step to generate a new keyphrase word by word. Both PD and WD repeat until meeting the stop conditions. In our method, both PD and WD attend the document content to gather contextual information. Moreover, the attention score of each WD step is rescaled by the corresponding PD attention score. The purpose of the attention rescaling is to indicate which aspect is focused on by the current PD step.

We also propose two kinds of exclusion mechanisms (i.e., a soft one and a hard one) to avoid generating duplicated keyphrases. Either the soft one or the hard one is used in our hierarchical decoding process.
Both of them are used in the WD process of our hierarchical decoding. Besides, both of them collect the previously-generated $K$ keyphrases, where $K$ is a predefined window size. The soft exclusion mechanism is incorporated in the training stage, where an exclusive loss is employed to encourage the model to generate a different first word of the current keyphrase with the first words of the collected $K$ keyphrases. However, the hard exclusion mechanism is used in the inference stage, where an exclusive search is used to force WD to produce a different first word with the first words of the collected $K$ keyphrases.
Our motivation is from the statistical observation that in 85\% of the documents on the largest KG benchmark, the keyphrases of each individual document have different first words. Moreover, since a keyphrase is usually composed of only two or three words, the predicted first word significantly affects the prediction of the following keyphrase words. Thus, our exclusion mechanisms can boost the diversity of the generated keyphrases. In addition, generating fewer duplications will also improve the chance to produce correct keyphrases that have not been predicted yet. 

We conduct extensive experiments on four popular real-world benchmarks. Empirical results demonstrate the effectiveness of our hierarchical decoding process. Besides, both the soft and the hard exclusion mechanisms significantly reduce the number of duplicated keyphrases. Furthermore, after employing the hard exclusion mechanism, our model consistently outperforms all the SOTA sequential decoding baselines on the four benchmarks.

We summarize our main contributions as follows: (1) to our best knowledge, we are the first to design a hierarchical decoding process for the keyphrase generation problem; (2) we propose two novel exclusion mechanisms to avoid generating duplicated keyphrases as well as improve the generation accuracy; and (3) our method consistently outperforms all the SOTA sequential decoding methods on multiple benchmarks under the new setting. 

\section{Related Work}
\subsection{Keyphrase Extraction}
Most of the traditional extractive methods~\cite{witten2005kea,mihalcea2004textrank} focus on extracting present keyphrases from the input document and follow a two-step framework. They first extract plenty of keyphrase candidates by handcrafted rules~\cite{medelyan2009human_maui}. Then, they score and rank these candidates based on either unsupervised methods~\cite{mihalcea2004textrank} or supervised learning methods~\cite{Nguyen2007NUS,Hulth2003inspec}. Recently, neural-based sequence labeling methods~\cite{gollapalli2017incorporating_seqlabel,luan2017scientific_seqlabel,zhang2016keyphrase_seqlabel_twitter} are also explored in keyphrase extraction problem. 
However, these extractive methods cannot predict absent keyphrase which is also an essential part of a keyphrase set.

\subsection{Keyphrase Generation}
To produce both present and absent keyphrases, \citet{Meng2017dkg} introduced a generative model, CopyRNN, which is based on an attentional encoder-decoder framework~\cite{DBLP:conf/iclr/BahdanauCB14} incorporating with a copy mechanism~\cite{Gu2016copy}. A wide range of extensions of CopyRNN are recently proposed~\cite{Chen2018corr_dkg,Chen2018TG_net,Ye2018semi_dkg,Chen2019Integrated,Zhao2019Linguistic_dkg}. All of them rely on beam search to over-generate lots of keyphrases with large beam size and then select the top $N$ (e.g., five or ten) ranked ones as the final prediction. That means these over-generated methods will always predict $N$ keyphrases for any input documents. Nevertheless, in a real situation, the keyphrase number should be determined by the document content and may vary among different documents. 

To this end, \citet{yuan2018generating_diverse} introduced a new setting that the KG model should predict multiple keyphrases and simultaneously decide the suitable keyphrase number for the given document. Two models with a sequential decoding process, catSeq and catSeqD, are proposed in~\citet{yuan2018generating_diverse}. The catSeq is also an attentional encoder-decoder model~\cite{DBLP:conf/iclr/BahdanauCB14} with copy mechanism~\cite{see2017get}, but adopting new training and inference setup to fit the new setting. The catSeqD is an extension of catSeq with orthogonal regularization~\cite{DBLP:conf/nips/BousmalisTSKE16_orthog} and target encoding. Lately, \citet{ken2019rl_dkg} proposed a reinforcement learning based fine-tuning method, which fine-tunes the pre-trained models with adaptive rewards for generating more sufficient and accurate keyphrases.
We follow the same setting with~\citet{yuan2018generating_diverse} and propose an exclusive hierarchical decoding method for the KG problem. To the best of our knowledge, this is the first time the hierarchical decoding is explored in the KG problem. Different from the hierarchical decoding in other areas~\cite{Fan2018HRD_story,Yarats2018HRD_dialogue,Tan2017HRD_summarization,Chen2018HRD_summarization}, we rescale the attention score of each WD step with the corresponding PD attention score to provide aspect guidance when generating keyphrases. Moreover, either a soft or a hard exclusion mechanism is innovatively incorporated in the decoding process to improve generation diversity.

\begin{figure*}[t]
\centering
\includegraphics[width=\textwidth]{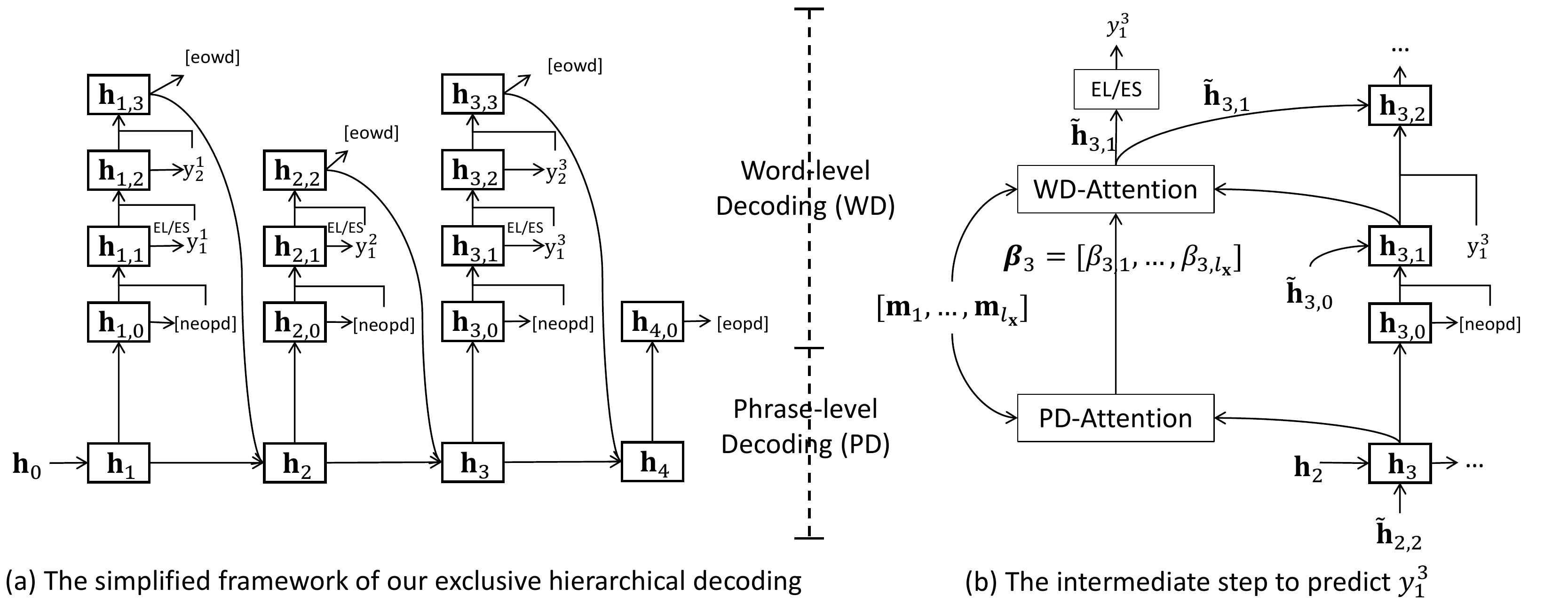}
\caption{Illustration of our exclusive hierarchical decoding. $\mathbf{h}_i$ is the hidden state of $i$-th PD step. $\mathbf{h}_{i,j}$ is the corresponding $j$-th WD hidden state. The ``[neopd]'' token means PD does not end. The ``[eowd]'' token means WD terminates. The ``[eopd]'' token means PD ends and the whole decoding process finishes. ``[$\mathbf{m}_1, \dots, \mathbf{m}_{l_{\mathbf{x}}}$]'' represents the encoded hidden states from the document. ``PD-Attention'' and ``WD-Attention'' are the attention mechanisms in PD and WD respectively. ``$\boldsymbol{\beta}_i$'' is the PD attention score at $i$-th step. ${\mathbf{\tilde{h}}_{i,j}}$ is the WD attentional vector. ``EL/ES'' indicates either the exclusive loss or the exclusive search is incorporated. 
}
\label{figure: model_seqE2_HRDv2}
\end{figure*}

\section{Notations and Problem Definition}
We denote vectors and matrices with bold lowercase and uppercase letters respectively. Sets are denoted with calligraphy letters. We use $\mathbf{W}$ to represent a parameter matrix. 

We define the keyphrase generation problem as follows. The input is a document $\mathbf{x}$, the output is a keyphrase set $\mathcal{Y}=\{\mathbf{y}^i\}_{i=1,\dots,|\mathcal{Y}|}$, where $|\mathcal{Y}|$ is the keyphrase number of $\mathbf{x}$. Both the $\mathbf{x}$ and each $\mathbf{y}^i$ are sequences of words, i.e., $\mathbf{x}=[x_1, ..., x_{l_\mathbf{x}}]$ and $\mathbf{y}^i=[y^i_1, ..., y_{l_{\mathbf{y}^i}}^i]$, where $l_\mathbf{x}$ and $l_{\mathbf{y}^i}$ are the word numbers of $\mathbf{x}$ and $\mathbf{y}^i$ correspondingly.

\section{Our Methodology}
We first encode each word of the document into a hidden state and then employ our exclusive hierarchical decoding shown in Figure~\ref{figure: model_seqE2_HRDv2} to produce keyphrases for the given document. Our hierarchical decoding process consists of phrase-level decoding (PD) and word-level decoding (WD). Each PD step decides an appropriate aspect to summarize based on both the context of the document and the aspects summarized by previous PD steps. Then, the hidden representation of the captured aspect is employed to initialize the WD process to generate a new keyphrase word by word. The WD process terminates when producing a ``[eowd]'' token. If the WD process output a ``[eopd]'' token, the whole hierarchical decoding process stops. Both PD and WD attend the document content. The PD attention score is used to re-weight the WD attention score to provide aspect guidance. To improve the diversity of the predicted keyphrases, we incorporate either an exclusive loss when training (i.e., the soft exclusion mechanism) or an exclusive search mechanism when inference (i.e., the hard exclusion mechanism). 

\subsection{Sequential Encoder}
To obtain the context-aware representation of each document word, we employ a two-layered bidirectional GRU~\cite{cho2014gru} as the document encoder: $\mathbf{m}_k = \text{BiGRU}(\mathbf{e}_{x_k}, \overrightarrow{\mathbf{m}}_{k-1},\overleftarrow{\mathbf{m}}_{k+1}),$
where $k=1,2,...,l_{\mathbf{x}}$ and $\mathbf{e}_{x_k}$ is the embedding vector of $x_k$ with $d_e$ dimensions. $\mathbf{m}_{k}=[\overrightarrow{\mathbf{m}}_{k}; \overleftarrow{\mathbf{m}}_{k}] \in \mathbb{R}^{d}$ is the encoded context-aware representation of $x_k$. Here, ``[$\cdot$ ; $\cdot$]'' means concatenation. 

\subsection{Hierarchical Decoder}
Our hierarchical decoding process is controlled by the hierarchical decoder, which utilizes a phrase-level decoder and a word-level decoder to handle the PD process and the WD process respectively. We present our hierarchical decoder first and then introduce the exclusion mechanisms. In our decoders, all the hidden states and attentional vectors are $d$-dimensional vectors. 

\subsubsection{Phrase-level Decoder} We adopt a unidirectional GRU layer as our phrase-level decoder. After the WD process under last PD step is finished, the phrase-level decoder will update its hidden state as follows: 
\begin{align}
\mathbf{h}_i=\overrightarrow{\text{GRU}}_1(\tilde{\mathbf{h}}_{i-1,end},\mathbf{h}_{i-1}),
\end{align}
where $\tilde{\mathbf{h}}_{i-1,end}$ is the attentional vector for the ending WD step under the ($i$-1)-th PD step (e.g., $\tilde{\mathbf{h}}_{2,2}$ in Figure~\ref{figure: model_seqE2_HRDv2}(b)). $\mathbf{h}_i$ is regarded as the hidden representation of the captured aspect at the $i$-th PD step.
$\mathbf{h}_0$ is initialized as the document representation $[\overrightarrow{\mathbf{m}}_{l_{\mathbf{x}}}; \overleftarrow{\mathbf{m}}_{1}]$. $\tilde{\mathbf{h}}_{0,end}$ is initialized with zeros. 

In PD-Attention process, the PD attentional score $\boldsymbol{\beta}_{i} = [\beta_{i,1},\beta_{i,2},\ldots,\beta_{i,l_{\mathbf{x}}}]$ is computed from the following attention mechanism employing $\mathbf{h}_i$ as the query vector:
\begin{align}
    \beta_{i,k} &= \exp(s_{i,k}) \slash \sum_{n=1}^{l_{\mathbf{x}}} \exp(s_{i,n}),\\
    s_{i,n} &= (\mathbf{h}_i)^T \mathbf{W}_1 \mathbf{m}_n.
\end{align}

\subsubsection{Word-level Decoder} We choose another unidirectional GRU layer to conduct word-level decoding. Under the $i$-th PD step, the word-level decoder updates its hidden state first:
\begin{align}
   \mathbf{h}_{i,j} = \overrightarrow{\text{GRU}}_2([\tilde{\mathbf{h}}_{i, j-1}; \mathbf{e}_{y^i_{j-1}}], \mathbf{h}_{i,j-1}),
\end{align}
where $\tilde{\mathbf{h}}_{i, j-1}$ is the WD attentional vector of the ($j$-1)-th WD step and $\mathbf{e}_{y^i_{j-1}}$ is the $d_e$-dimensional embedding vector of the $y^i_{j-1}$ token. We define $\mathbf{h}_{i,0}=\overrightarrow{\text{GRU}}_2([\mathbf{0}; \mathbf{e}_{s}], \mathbf{h}_{i})$, where $\mathbf{h}_i$ is the current hidden state of the phrase-level decoder, $\mathbf{0}$ is a zero vector, and $\mathbf{e}_s$ is the embedding of the start token.
Then, the WD attentional vector is computed: 
\begin{align}
    \tilde{\mathbf{h}}_{i,j} &= \text{tanh}(\mathbf{W}_2[\mathbf{h}_{i,j};\mathbf{a}_{i,j}]),\\
    \mathbf{a}_{i,j} &=  \sum_{k=1}^{l_{\mathbf{x}}} \bar{\alpha}_{(i,j),k} \mathbf{m}_k, \\
    \bar{\alpha}_{(i,j),k} &= \frac{\alpha_{(i,j),k}\times\beta_{i,k}}{\sum_{n=1}^{l_{\mathbf{x}}}\alpha_{(i,j),n}\times\beta_{i,n}} \label{eq: hr_rescale},
\end{align}
where $\alpha_{(i,j),k}$ is the original WD attention score which is computed similar to $\beta_{i,k}$ except that a new parameter matrix is used and $\mathbf{h}_{i,j}$ is employed as the query vector.
The purpose of the rescaling operation in Eq.~(\ref{eq: hr_rescale}) is to indicate the focused aspect of the current PD step for each WD step.

Finally, the $\tilde{\mathbf{h}}_{i,j}$ is utilized to predict the probability distribution of current keyword with the copy mechanism~\cite{see2017get}:
\begin{align}
    P^i_j &= (1-g^i_j)P^i_{j, \mathcal{V}} + g^i_j P^i_{j, \mathcal{X}},
\end{align}
where $g^i_j = \text{sigmoid}(\mathbf{w}^T_g \tilde{\mathbf{h}}_{i,j} + b_g) \in \mathbb{R}$ is the copy gate. $P^i_{j, \mathcal{V}} = \text{softmax}(\mathbf{W}_3 \tilde{\mathbf{h}}_{i,j} + \mathbf{b}_{\mathcal{V}}) \in \mathbb{R}^{|\mathcal{V}|}$ is the probability distribution over a predefined vocabulary $\mathcal{V}$. $P^i_{j, \mathcal{X}} = \sum_{k:x_k=y^i_j} \bar{\alpha}_{(i,j),k} \in \mathbb{R}^{|\mathcal{X}|}$ is the copying probability distribution over $\mathcal{X}$ which is a set of all the words that appeared in the document. $P^i_j \in \mathbb{R}^{|\mathcal{V} \cup \mathcal{X}|}$ is the final predicted probability distribution. Finally, greedy search is applied to produce the current token.

The WD process terminates when producing a ``[eowd]'' token. The whole hierarchical decoding process ends if the word-level decoder produces a ``[eopd]'' token at the $0$-th step, i.e., $y^i_0$ is predicted as ``[eopd]''.

\subsection{Training}
A standard negative log-likelihood loss is employed as the generation loss to train our hierarchical decoding model:
\begin{equation}
    \mathcal{L}_{g} = -\sum_{i=1}^{|\mathcal{\bar{Y}}|}\sum_{j=0}^{l_{\mathbf{\bar{y}}^{i}}}\log P^i_j(\bar{y}^{i}_j|\mathbf{x};\mathbf{\bar{Y}}^{i-1}; \bar{\mathbf{y}}^{i}_{j-1}),
\end{equation}
where $\mathbf{\bar{Y}}^{i-1}=\mathbf{\bar{y}}^{1}, \dots, \mathbf{\bar{y}}^{i-1}$ are the target keyphrases of previously-finished PD steps and $\bar{\mathbf{y}}^{i}_{j-1} = \bar{y}^i_0, \dots, \bar{y}^i_{j-1}$ are target keyphrase words of previous WD steps under the $i$-th PD step. When training, each original target keyphrase is extended with a ``[neopd]'' token and a ``[eowd]'' token, i.e., $\mathbf{\bar{y}}^i=[\text{``[neopd]''}, y^i_1, \dots, y^i_{l_{\mathbf{y}^i}}, \text{``[eowd]''}]$. Besides, a ``[eopd]'' token is also incorporated into the targets to indicate the ending of whole decoding process. Teacher forcing is employed when training.

\subsection{Soft and Hard Exclusion Mechanisms}
To alleviate the duplication generation problem, we propose a soft and a hard exclusion mechanisms. Either of them can be incorporated into our hierarchical decoding process to form one kind of exclusive hierarchical decoding method.

\smallskip
\noindent \textbf{Soft Exclusion Mechanism}. An exclusive loss (EL) is introduced in the training stage as shown in Algorithm~\ref{alg: exclusive loss}. ``$j == 1$'' in line ``3'' means the current WD step is predicting the first word of a keyphrase. In short, the exclusive loss punishes the model for the tendency to generate the same first word of the current keyphrase with the first words of previously-generated keyphrases within the window size $K_{EL}$.

\smallskip
\noindent \textbf{Hard Exclusion Mechanism}. An exclusive search (ES) is introduced in the inference stage as shown in Algorithm~\ref{alg: exclusive search}. The exclusive search mechanism forces the word-level decoding to predict a different first word with the first words of previously-predicted keyphrases within the window size $K_{ES}$. 

Since a keyphrase usually has only two or three words, the first word significantly affects the prediction of the following words. Therefore, both the soft and the hard exclusion mechanisms can improve the diversity of generated keyphrases.

\begin{algorithm}[t]
\small
\caption{Training with Exclusive Loss}
\label{alg: exclusive loss}
\begin{algorithmic}[1]
\REQUIRE The window size $K_{EL}$. The target keyphrases $[\mathbf{\bar{y}}^1, \dots, \mathbf{\bar{y}}^{i}, \dots, \mathbf{\bar{y}^{|\mathcal{\bar{Y}}|}}]$. The predicted probability distribution $P^i_j$ for the $j$-th WD step under the $i$-th PD step where $i=1, \dots, |\mathcal{\bar{Y}}|$ and $j=0, 1, \dots, l_{\mathbf{\bar{y}}^{i}}$.
\STATE Firstly, the exclusive loss of the $j$-th WD step under the $i$-th PD step is computed as follows.
\STATE $K_{EL} \gets \text{min}\{K_{EL}, i-1\}$
\IF {$K_{EL} > 0$ \AND $j == 1$}
    \STATE $\mathcal{L}_{EL}^{i,j} = \sum_{idx = i-K_{EL}, \bar{y}^{idx}_j \ne \bar{y}^i_j}^{i-1} - \log (1 - P^i_j(\bar{y}^{idx}_j))$
\ELSE
    \STATE $\mathcal{L}_{EL}^{i,j} = 0.0$
\ENDIF
\STATE Secondly, the exclusive loss for the whole decoding process is calculated as $\mathcal{L}_{EL} = \sum_{i,j} \mathcal{L}_{EL}^{i,j}$.
\STATE Finally, the joint loss $\mathcal{L} = \mathcal{L}_{g} + \mathcal{L}_{EL}$ is employed to train the model.
\end{algorithmic}
\end{algorithm}

\begin{algorithm}[t]
\small
\caption{Inference with Exclusive Search}
\label{alg: exclusive search}
\begin{algorithmic}[1]
\REQUIRE The window size $K_{ES}$. The first words of previously-predicted keyphrases $[y^1_1, \dots, y^{i-1}_1]$. The current WD step index $j$. The predicted probability distribution $P^i_j$ for current WD step.  
\STATE $K_{ES} \gets \text{min}\{K_{ES}, i-1\}$
\IF {$K_{ES} > 0$ \AND $j == 1$}
    \FOR {$idx = i-K_{ES}, i-K_{ES}+1, \dots, i-1$}
        \STATE $P^i_j(y^{idx}_j) \gets 0.0$
    \ENDFOR
\ENDIF
\STATE Return $y^i_j=\arg \max(P^i_j)$ as the predicted word for current WD step.
\end{algorithmic}
\end{algorithm}

\section{Experiment Setup}
Our model implementations are based on the OpenNMT system~\cite{klein2017opennmt} using PyTorch~\cite{paszke2017automatic_pytorch}. Experiments of all models are repeated with three different random seeds and the averaged results are reported.
\subsection{Datasets}
We employ four scientific article benchmark datasets to evaluate our models, including \textbf{KP20k}~\cite{Meng2017dkg}, \textbf{Inspec}~\cite{Hulth2003inspec}, \textbf{Krapivin}~\cite{Krapivin2009LargeDF}, and \textbf{SemEval}~\cite{Kim2010SemEval}. Following previous work~\cite{yuan2018generating_diverse,Chen2019Integrated}, we use the training set of KP20k to train all the models. After removing the duplicated data, we maintain 509,818 data samples in the training set, 20,000 in the validation set, and 20,000 in the testing set. After training, we test all the models on the testing datasets of these four benchmarks. The dataset statistics are shown in Table~\ref{table:statistic_testset}.

\begin{table}[ht]
\centering
\begin{tabular}{ |c| c| c| c|}
\hline
 Dataset & \textbf{Total} & \textbf{Validation} & \textbf{Testing}\\
\hline
Inspec & 2,000 & 1,500 & 500\\
\hline
Krapivin & 2,303 & 1,903 & 400\\
\hline
SemEval & 244 & 144 & 100\\
\hline
KP20k & 549,818 & 20,000 & 20,000\\
\hline
\end{tabular}
\caption{The statistics of validation and testing datasets.}
\label{table:statistic_testset}
\end{table}

\subsection{Baselines}
We focus on the comparisons with state-of-the-art decoding methods and choose the following generation models under the new setting as our baselines:
\begin{itemize}[leftmargin=*]
    \item \textbf{Transformer}~\cite{DBLP:conf/nips/VaswaniSPUJGKP17_transformer}. A transformer-based sequence to sequence model incorporating with copy mechanism.
    
    \item \textbf{catSeq}~\cite{yuan2018generating_diverse}. An RNN-based attentional encoder-decoder model with copy mechanism. Both the encoding and decoding are sequential.
    
    \item \textbf{catSeqD}~\cite{yuan2018generating_diverse}. An extension of catSeq which incorporates orthogonal regularization~\cite{DBLP:conf/nips/BousmalisTSKE16_orthog} and target encoding into the sequential decoding process to improve the generation diversity and accuracy.
    
    \item \textbf{catSeqCorr}~\cite{ken2019rl_dkg}. Another extension of catSeq, which incorporates the sequential decoding with coverage~\cite{see2017get} and review mechanisms  to boost the generation diversity and accuracy. This method is adjusted from~\citet{Chen2018corr_dkg} to fit the new setting.
\end{itemize}

In this paper, we propose two novel models that are denoted as follows:
\begin{itemize}[leftmargin=*]
    \item \textbf{ExHiRD-s}. Our \textbf{Ex}clusive \textbf{Hi}e\textbf{R}archical \textbf{D}ecoding model with the \textbf{s}oft exclusion mechanism. In experiments, the window size $K_{EL}$ is selected as 4 after tuning on the KP20k validation dataset.
    \item \textbf{ExHiRD-h}. Our \textbf{Ex}clusive \textbf{Hi}e\textbf{R}archical \textbf{D}ecoding model with the \textbf{h}ard exclusion mechanism. In experiments, the values of the window size $K_{ES}$ are selected as 4, 1, 1, 1 for Inspec, Krapivin, SemEval, and KP20k respectively after tuning on the corresponding validation datasets.
\end{itemize}

We choose the bilinear attention from~\citet{luong2015effective} and the copy mechanism from~\citet{see2017get} for all the models.

\begin{table*}[t]
\centering
\small
\resizebox{0.78\textwidth}{!}{
\begin{tabular}{ l | c c | c c | c c | c c}
\hline \hline
\multicolumn{1}{c|}{\multirow{2}{*}{\textbf{Model}}} & \multicolumn{2}{c|}{\textbf{Inspec}} & \multicolumn{2}{c|}{\textbf{Krapivin}} & \multicolumn{2}{c|}{\textbf{SemEval}} & \multicolumn{2}{c}{\textbf{KP20k}} \\
\multicolumn{1}{c|}{}   & $F_{1}@M$    & $F_{1}@5$    & $F_{1}@M$  & $F_{1}@5$ & $F_{1}@M$    & $F_{1}@5$   & $F_{1}@M$   & $F_{1}@5$  \\
\hline \hline
Transformer & 0.254\textsubscript{5} & 0.210\textsubscript{7} & 0.328\textsubscript{14} & 0.252\textsubscript{4} & 0.310\textsubscript{5} & 0.257\textsubscript{4} & 0.360\textsubscript{3} & 0.282\textsubscript{10} \\
catSeq  & 0.276\textsubscript{5} & 0.233\textsubscript{4} & 0.344\textsubscript{14} & 0.269\textsubscript{5} & 0.313\textsubscript{8} & 0.262\textsubscript{11} & 0.368\textsubscript{1} & 0.295\textsubscript{2}   \\
catSeqD & 0.280\textsubscript{3} & 0.236\textsubscript{1} & 0.344\textsubscript{9} & 0.268\textsubscript{8} & 0.311\textsubscript{6} & 0.263\textsubscript{6} & 0.368\textsubscript{2} & 0.296\textsubscript{2}    \\
catSeqCorr & 0.253\textsubscript{3} & 0.208\textsubscript{6} & 0.343\textsubscript{9} & 0.258\textsubscript{9} & 0.318\textsubscript{18} & 0.260\textsubscript{14} & 0.367\textsubscript{3} & 0.281\textsubscript{4} \\
\hline
ExHiRD-s & 0.278\textsubscript{5} & 0.235\textsubscript{3} & 0.338\textsubscript{3} & 0.278\textsubscript{0} & 0.322\textsubscript{5} & 0.276\textsubscript{5} & 0.372\textsubscript{1} & 0.307\textsubscript{0} \\
ExHiRD-h & \textbf{0.291$_\text{3}$} & \textbf{0.253$_\text{4}$} & \textbf{0.347\textsubscript{4}} & \textbf{0.286$_\text{4}$} & \textbf{0.335\textsubscript{17}} & \textbf{0.284\textsubscript{15}} & \textbf{0.374$_\text{0}$} & \textbf{0.311$_\text{1}$} \\
\hline
\end{tabular}
}
\caption{
Present keyphrase prediction results of all models on all datasets. The best results are bold. In all the tables of this paper, the subscript represents the corresponding standard deviation (e.g., 0.311\textsubscript{1} indicates 0.311$\pm$0.001).
}
\label{table:present-result}
\end{table*}

\begin{table*}[t]
\centering
\small
\resizebox{0.78\textwidth}{!}{
\begin{tabular}{ l | c c | c c | c c | c c}
\hline \hline
\multicolumn{1}{c|}{\multirow{2}{*}{\textbf{Model}}} & \multicolumn{2}{c|}{\textbf{Inspec}} & \multicolumn{2}{c|}{\textbf{Krapivin}} & \multicolumn{2}{c|}{\textbf{SemEval}} & \multicolumn{2}{c}{\textbf{KP20k}} \\
\multicolumn{1}{c|}{} & $F_{1}@M$    & $F_{1}@5$    & $F_{1}@M$  & $F_{1}@5$ & $F_{1}@M$    & $F_{1}@5$   & $F_{1}@M$   & $F_{1}@5$  \\
\hline \hline
Transformer & 0.013\textsubscript{1} & 0.006\textsubscript{1} & 0.030\textsubscript{5} & 0.014\textsubscript{3} & 0.020\textsubscript{1} & 0.013\textsubscript{1}& 0.024\textsubscript{2} & 0.011\textsubscript{1} \\
catSeq  & 0.008\textsubscript{3} & 0.004\textsubscript{1} 
        & 0.033\textsubscript{4} & 0.015\textsubscript{2} 
        & 0.017\textsubscript{2} & 0.012\textsubscript{1} 
        & 0.023\textsubscript{1} & 0.010\textsubscript{0} \\
catSeqD & 0.010\textsubscript{4} & 0.004\textsubscript{1} 
        & 0.033\textsubscript{7} & 0.015\textsubscript{3} 
        & 0.016\textsubscript{1} & 0.011\textsubscript{1} 
        & 0.023\textsubscript{1} & 0.010\textsubscript{1} \\
catSeqCorr & 0.007\textsubscript{2} & 0.004\textsubscript{1} 
           & 0.022\textsubscript{6} & 0.011\textsubscript{3} 
           & 0.021\textsubscript{5} & 0.014\textsubscript{3} 
           & 0.023\textsubscript{1} & 0.010\textsubscript{1} \\
\hline
ExHiRD-s & 0.021\textsubscript{7} & 0.009\textsubscript{2} 
         & 0.033\textsubscript{5} & 0.016\textsubscript{2}
         & 0.024\textsubscript{5} & 0.016\textsubscript{4}
         & 0.029\textsubscript{1} & 0.014\textsubscript{0} \\
ExHiRD-h    & \textbf{0.022$_\text{3}$} & \textbf{0.011$_\text{1}$} 
         & \textbf{0.043$_\text{6}$} & \textbf{0.022$_\text{3}$} 
         & \textbf{0.025\textsubscript{6}} & \textbf{0.017\textsubscript{4}} 
         & \textbf{0.032$_\text{0}$} & \textbf{0.016$_\text{0}$} \\
\hline
\end{tabular}
}
\caption{
Absent keyphrase prediction results of all models on all datasets. The best results are bold. 
}
\label{table:absent-result}
\end{table*}

\subsection{Evaluation Metrics}
We engage $F_1@M$ which is recently proposed in~\citet{yuan2018generating_diverse} as one of our evaluation metrics. $F_1@M$ compares all the predicted keyphrases by the model with ground-truth keyphrases, which means it does not use a fixed cutoff for the predictions. Therefore, it considers the number of predictions.

We also use $F_1@5$ as another evaluation metric. When the number of predictions is less than five, we randomly append incorrect keyphrases until it obtains five predictions instead of directly using the original predictions. If we do not adopt such an appending operation, $F_1@5$ will become the same with $F_1@M$ when the prediction number is less than five.

The macro-averaged $F_1@M$ and $F_1@5$ scores are reported. When determining whether two keyphrases are identical, all the keyphrases are stemmed first. Besides, all the duplicated keyphrases are removed after stemming.

\subsection{Implementation Details}
Following previous work~\cite{Meng2017dkg,yuan2018generating_diverse,Chen2019Integrated,ken2019rl_dkg}, we lowercase the characters, tokenize the sequences, and replace digits with ``$<$digit$>$'' token. Similar to~\citet{yuan2018generating_diverse}, when training, the present keyphrase targets are sorted according to the orders of their first occurrences in the document. Then, the absent keyphrase targets are put at the end of the sorted present keyphrase targets. We use ``$<$p\_start$>$'' and ``$<$a\_start$>$'' as the ``[neopd]'' token of present and absent keyphrases respectively. ``;'' is employed as the ``[eowd]'' token for both present and absent keyphrases. ``$<$/s$>$'' is used as the ``[eopd]'' token.

The vocabulary with 50,000 tokens is shared between the encoder and decoder. We set $d_e$ as 100 and $d$ as 300. The hidden states of the encoder layers are initialized as zeros. In the training stage, we randomly initialize all the trainable parameters including the embedding using a uniform distribution in $[-0.1, 0.1]$. We set batch size as 10, max gradient norm as 1.0, and initial learning rate as 0.001. We do not use dropout. Adam~\cite{Kingma2014Adam} is used as our optimizer. 
The learning rate decays to half if the perplexity on KP20k validation set stops decreasing. Early stopping is applied when training. 
When inference, we set the minimum phrase-level decoding step as 1 and the maximum as 20.

\section{Results and Analysis}
\subsection{Present and Absent Keyphrase Predictions}
We show the present and absent keyphrase prediction results in Table~\ref{table:present-result} and Table~\ref{table:absent-result} correspondingly.
As indicated in these two tables, both the ExHiRD-s model and the ExHiRD-h outperform the state-of-the-art baselines on most of the metrics, which demonstrates the effectiveness of our exclusive hierarchical decoding methods. Besides, the ExHiRD-h model consistently achieves the best results on both present and absent keyphrase prediction in all the datasets\footnote{We also tried to simultaneously incorporate the soft and the hard exclusion mechanisms into our hierarchical decoding model, but it still underperforms ExHiRD-h.}.

\begin{table}[t]
\centering
\resizebox{0.98\columnwidth}{!}{
\begin{tabular}{l| c| c| c| c}
  \hline
  \hline
  \textbf{Model} & \textbf{Inspec} & \textbf{Krapivin} & \textbf{SemEval} & \textbf{KP20k}\\
  \hline
  \hline
  Transformer & 0.286\textsubscript{25} & 0.297\textsubscript{46} & 0.220\textsubscript{38} & 0.223\textsubscript{41} \\
  catSeq    & 0.302\textsubscript{11}  & 0.277\textsubscript{8} & 0.200\textsubscript{2}  & 0.217\textsubscript{4}\\
 catSeqD    & 0.304\textsubscript{14} & 0.283\textsubscript{9}  & 0.199\textsubscript{1}  & 0.215\textsubscript{8}\\
 catSeqCorr & 0.352\textsubscript{38} & 0.354\textsubscript{4} & 0.249\textsubscript{23} & 0.282\textsubscript{14}\\
 \hline
 ExHiRD-s   & 0.210\textsubscript{14} & 0.182\textsubscript{12} & 0.119\textsubscript{8} & 0.137\textsubscript{6}\\
  ExHiRD-h    & \textbf{0.030$_\text{6}$}  & \textbf{0.140$_\text{6}$} & \textbf{0.091$_\text{10}$} & \textbf{0.110$_\text{1}$}\\
  \hline
\end{tabular}
}
\caption{The average DupRatios of predicted keyphrases on all datasets. The lower the score, the better the performance. 
}
\label{table:dupratios of predicted keyphrases}
\end{table}

\subsection{Duplication Ratio of Predicted Keyphrases}
In this section, we study the model capability of avoiding producing duplicated keyphrases. Duplication ratio is denoted as ``DupRatio'' and defined as follows:
\begin{align}
    DupRatio = \frac{\text{\#}~duplications}{\text{\#}~ predictions},
\end{align}
where \# means ``the number of''. For instance, the DupRatio is 0.5 (3/6) for [A, A, B, B, A, C]. 

We report the average DupRatio per document in Table~\ref{table:dupratios of predicted keyphrases}. From this table, we observe that our ExHiRD-s and ExHiRD-h consistently and significantly reduce the duplication ratios on all datasets. Moreover, we also find that our ExHiRD-h model achieves the lowest duplication ratios on all datasets.

\begin{table}[t]
\centering
\resizebox{0.98\columnwidth}{!}{
\begin{tabular}{l| c c| c c| c c| c c}
  \hline
  \hline
  \multirow{2}{*}{\textbf{Model}} & \multicolumn{2}{c|}{\textbf{Inspec}} & \multicolumn{2}{c|}{\textbf{Krapivin}} &
  \multicolumn{2}{c|}{\textbf{SemEval}} & \multicolumn{2}{c}{\textbf{KP20k}}
  \\
  & \text{\#}PK & \text{\#}AK & \text{\#}PK & \text{\#}AK & \text{\#}PK & \text{\#}AK & \text{\#}PK & \text{\#}AK\\
  \hline
  \hline
 Oracle        & 7.64 & 2.10 & 3.27 & 2.57 & 6.28 & 8.12 & 3.32 & 1.93\\
 \hline
 Transformer & 3.17\textsubscript{10} & 0.70\textsubscript{4} & 3.57\textsubscript{29} & 0.63\textsubscript{4} & 3.24\textsubscript{20} & 0.67\textsubscript{3}  & 3.44\textsubscript{17} & 0.58\textsubscript{4}\\
 catSeq        & 3.33\textsubscript{2} & 0.58\textsubscript{4} & 3.70\textsubscript{10} & 0.63\textsubscript{5} & 3.45\textsubscript{5} & 0.64\textsubscript{3} & 3.70\textsubscript{4} & 0.51\textsubscript{2}\\
 catSeqD       & 3.33\textsubscript{4} & 0.58\textsubscript{2} & 3.66\textsubscript{10}  & 0.61\textsubscript{1} & 3.47\textsubscript{5} & 0.63\textsubscript{7} & 3.74\textsubscript{3} & 0.50\textsubscript{2}\\
 catSeqCorr    & 3.07\textsubscript{7} & 0.53\textsubscript{2} & \textbf{3.39\textsubscript{14}} & 0.56\textsubscript{1} & 3.15\textsubscript{3} & 0.62\textsubscript{1} & \textbf{3.36\textsubscript{4}} & 0.50\textsubscript{1}\\
 \hline
 ExHiRD-s      & 3.56\textsubscript{5} & 0.81\textsubscript{2} & 4.33\textsubscript{7} & 0.86\textsubscript{3}
               & \textbf{3.69\textsubscript{14}} & 0.79\textsubscript{6} & 3.94\textsubscript{2} & 0.69\textsubscript{1}\\
  ExHiRD-h        & \textbf{4.00\textsubscript{4}} & \textbf{1.50\textsubscript{6}} & 4.41\textsubscript{9} & \textbf{1.02\textsubscript{7}} & 3.65\textsubscript{13} & \textbf{0.99\textsubscript{4}} & 3.97\textsubscript{3} & \textbf{0.81\textsubscript{1}}\\
  \hline
\end{tabular}
}
\caption{Results of average numbers of predicted unique keyphrases per document. ``\text{\#}PK'' and ``\text{\#}AK'' are the number of present and absent keyphrases respectively.  ``Oracle'' is the gold average keyphrase number. The closest values to the oracles are bold. 
}
\label{table:number of predicted keyphrases}
\end{table}

\subsection{Number of Predicted Keyphrases}
We also study the average number of unique keyphrase predictions per document. Duplicated keyphrases are removed. The results are shown in Table~\ref{table:number of predicted keyphrases}. One main finding is that all the models generate an insufficient number of unique keyphrases on most datasets, especially for predicting absent keyphrases. We also observe that our methods can improve the number of unique keyphrases by a large margin, which is extremely beneficial to solve the problem of insufficient generation. Correspondingly, it also leads to over-generate more keyphrases than the ground-truth for the cases that do not have this problem, such as the present keyphrase predictions on Krapivin and KP20k datasets. We leave solving the over-generation of present keyphrases on Krapivin and KP20k as our future work.

\subsection{ExHiRD-h: Ablation Study}
Since our ExHiRD-h model achieves the best performance on almost all of the metrics, we select it as our final model and probe it more subtly in the following sections. In order to understand the effects of each component of ExHiRD-h, we conduct an ablation study on it and report the results on the SemEval dataset in Table~\ref{table:ablation_study_semeval}.

\begin{table}[t]
\centering
\resizebox{\columnwidth}{!}{
\begin{tabular}{l| c c c| c c c| c}
  \hline
  \hline
  \multirow{2}{*}{\textbf{Model}} & \multicolumn{3}{c|}{\textbf{Present}} & \multicolumn{3}{c|}{\textbf{Absent}} &
  \multirow{2}{*}{\textbf{DupRatio}}
  \\
  & $F_{1}@M$    & $F_{1}@5$ & \#PK & $F_{1}@M$   & $F_{1}@5$ & \#AK & \\
  \hline
  \hline
 ExHiRD-h      & \textbf{0.335} & \textbf{0.284} & \textbf{3.65} & \textbf{0.025} & \textbf{0.017} & \textbf{0.99} & \textbf{0.091} \\
 \hline
 w/o HRD       & 0.320 & 0.274 & 3.58 & 0.018 & 0.013 & 0.97 & 0.093   \\
 w/o ES        & 0.330 & 0.278 & 3.51 & 0.022 & 0.014 & 0.70 & 0.191   \\
  \hline
\end{tabular}
}
\caption{Ablation study of our ExHiRD-h model on SemEval dataset. ``w/o HRD'' means the hierarchical decoder is replaced with a sequential decoder and the exclusive search is still incorporated. ``w/o ES'' represents our hierarchical decoding model without utilizing exclusive search mechanism. 
}
\label{table:ablation_study_semeval}
\end{table}

We observe that both our hierarchical decoding process and exclusive search mechanism are helpful to generate more accurate present and absent keyphrases. Besides, we also find that the significant performance margins on the duplication ratio and the keyphrase numbers are mainly from the exclusive search mechanism. 

\begin{table}[t]
\centering
\resizebox{\columnwidth}{!}{
\begin{tabular}{c| c c c| c c c| c}
  \hline
  \hline
  \multirow{2}{*}{\textbf{$K_{ES}$}} & \multicolumn{3}{c|}{\textbf{Present}} & \multicolumn{3}{c|}{\textbf{Absent}} & \multirow{2}{*}{\textbf{DupRatio}}
  \\
  & $F_1@M$ & $F_1@5$ & \#PK & $F_1@M$ & $F_1@5$ & \#AK & \\
  \hline
  \hline
 Oracle & - & - & 3.32 & - & - & 1.93 & -\\
 \hline
 0   & 0.376 & 0.303 & 3.76 & 0.028 & 0.013 & 0.61 & 0.195 \\
 1   & 0.374 & 0.311 & 3.97 & 0.033 & 0.016 & 0.86 & 0.110\\
 2   & 0.371 & 0.314 & 4.11 & 0.034 & 0.017 & 1.00 & 0.069\\
 3   & 0.368 & 0.316 & 4.21 & 0.034 & 0.017 & 1.08 & 0.038\\
 4   & 0.366 & 0.316 & 4.27 & 0.033 & 0.017 & 1.16 & 0.017\\
 5   & 0.366 & 0.316 & 4.30 & 0.033 & 0.017 & 1.19 & 0.010\\
 all & 0.365 & 0.316 & 4.32 & 0.032 & 0.017 & 1.25 & 0.002\\
  \hline
\end{tabular}
}
\caption{Results of ExHiRD-h on KP20k with different window size $K_{ES}$. When $K_{ES}=0$, ExHiRD-h equals to ``w/o ES''. The ``all'' means we taking the first words of all the previously-predicted keyphrases into consideration. The ``DupRatio'' is the average DupRatio per document. We show the average numbers of ground-truth keyphrases in the ``Oracle'' row. 
}
\label{table:the study of window size}
\end{table}

\subsection{ExHiRD-h: Window Size of Exclusive Search}
For a more comprehensive understanding of our exclusive search mechanism in our ExHiRD-h model, we also study the effects of the window size $K_{ES}$. We conduct the experiments on KP20k dataset and list the results in Table~\ref{table:the study of window size}.

We note that a larger window size $K_{ES}$ leads to a lower DupRatio as we anticipated. It is because the exclusive search can observe more previously-generated keyphrases to avoid generating duplicated keyphrases when $K_{ES}$ is larger. When $K_{ES}$ is ``all'', the DupRatio is not absolute zero because we stem keyphrases when determining whether they are duplicated. Besides, we also find that larger $K_{ES}$ leads to better $F_1@5$ scores. The reason is that for $F_1@5$ scores, we append incorrect keyphrases to obtain five predictions when the number of predictions is less than five. A larger $K_{ES}$ leads to predict more unique keyphrases, append less absolutely incorrect keyphrases and improve the chance to output more accurate keyphrases.  However, generating more unique keyphrases may also lead to more incorrect predictions, which will degrade the $F_1@M$ scores since $F_1@M$ considers all the unique predictions without a fixed cutoff.

\begin{table}[t]
\centering
\resizebox{\columnwidth}{!}{
\begin{tabular}{l| c c c| c c c| c}
  \hline
  \hline
  \multirow{2}{*}{\textbf{Model}} & \multicolumn{3}{c|}{\textbf{Present}} & \multicolumn{3}{c|}{\textbf{Absent}} & \multirow{2}{*}{\textbf{DupRatio}}
  \\
  & $F_1@M$ & $F_1@5$ & \#PK & $F_1@M$ w& $F_1@5$ & \#AK & \\
  \hline
  \hline
 Oracle & - & - & 3.32 & - & - & 1.93 & -\\
 \hline
 Transformer      & 0.360 & 0.282 & 3.44 & 0.024 & 0.011 & 0.58 & 0.223 \\
 catSeq            & 0.368 & 0.295 & 3.70 & 0.023 & 0.010 & 0.51 & 0.217 \\
 catSeqD           & 0.368 & 0.296 & 3.74 & 0.023 & 0.010 & 0.50 & 0.215\\
 catSeqCorr        & 0.367 & 0.281 & \textbf{3.36} & 0.023 & 0.010 & 0.50 & 0.282\\
 \hline
 Transformer w/ ES      & 0.359 & 0.294 & 3.75 & 0.027 & 0.013 & 0.79 & 0.114\\
 catSeq w/ ES      & 0.366 & 0.305 & 3.95 & 0.025 & 0.012 & 0.68 & 0.138\\
 catSeqD w/ ES     & 0.366 & 0.306 & 3.99 & 0.026 & 0.012 & 0.65 & 0.137\\
 catSeqCorr w/ ES  & 0.366 & 0.298 & 3.74 & 0.027 & 0.013 & 0.72 & 0.159\\
 \hline
 ExHiRD-h          & \textbf{0.374} & \textbf{0.311} & 3.97 & \textbf{0.032} & \textbf{0.016} & \textbf{0.81} & \textbf{0.110}\\
  \hline
\end{tabular}
}
\caption{Results of applying our exclusive search to other baselines on KP20k. The ``w/ ES'' means our exclusive search is applied.
}
\label{table:ES on other baselines}
\end{table}

\subsection{ExHiRD-h: Incorporate Baselines with Exclusive Search}
Our exclusive search is a general method that can be easily applied to other models. In this section, we study the effects of our exclusive search on other baseline models. We show the experimental results on KP20k dataset in Table~\ref{table:ES on other baselines}.

From this table, we note that the effects of exclusive search on baselines are similar to the effects on our hierarchical decoding. We also see our ExHiRD-h still achieves the best performance on most of the metrics, even if baselines are also incorporated with exclusive search, which exhibits the superiority of our hierarchical decoding again.

\section{ExHiRD-h: Case Study}
\begin{figure}[t]
\centering
\includegraphics[width=0.98\columnwidth]{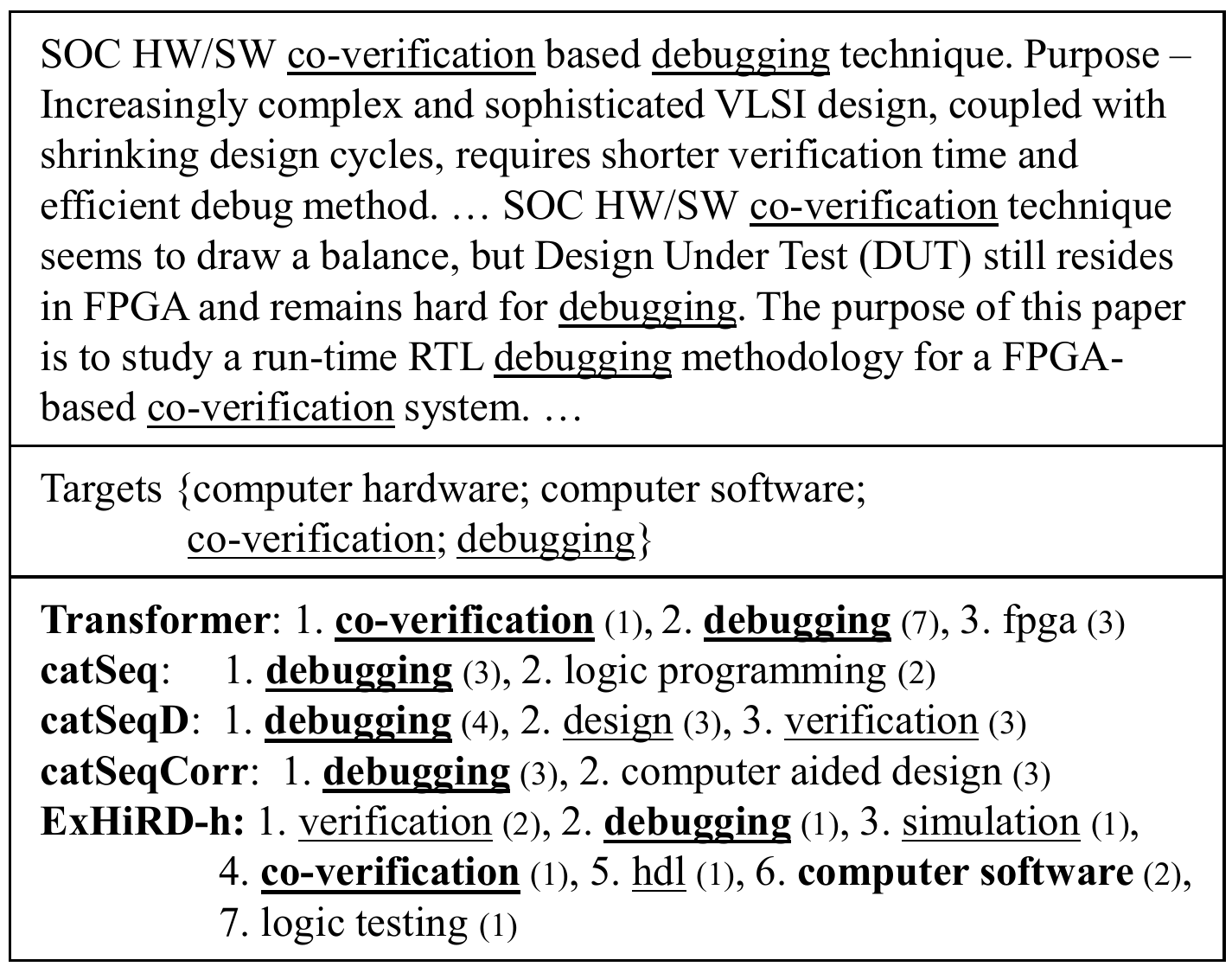}
\caption{An example of generated keyphrases by baselines and our ExHiRD-h. The correct predictions are bold and the present keyphrases are underlined. The digit in parentheses represents the frequency that the corresponding keyphrase is generated by the model (e.g., ``debugging (3)'' means the keyphrase ``debugging'' is generated three times by the model).}
\label{figure: case_study_final}
\end{figure}

We display a prediction example in Figure~\ref{figure: case_study_final}. Our ExHiRD-h model generates more accurate keyphrases for the document comparing to the four baselines. Besides, we also observe much less repeated keyphrases are generated by our ExHiRD-h. For instance, all the baselines produce the keyphrase ``debugging'' at least three times. However, our ExHiRD-h only generates it once, which demonstrates that our proposed method is more powerful in avoiding duplicated keyphrases. 

\section{Conclusion and Future Work}
In this paper, we propose an exclusive hierarchical decoding framework for keyphrase generation. Unlike previous sequential decoding methods, our hierarchical decoding consists of a phrase-level decoding process to capture the current aspect to summarize and a word-level decoding process to generate keyphrases based on the captured aspect. Besides, we also propose a soft and a hard exclusion mechanisms to enhance the diversity of the generated keyphrases. Extensive experimental results demonstrate the effectiveness of our methods. One interesting future direction is to explore whether the beam search is helpful to our model.

\section*{Acknowledgments}
The work described in this paper was partially supported by the Research Grants Council of the Hong Kong Special Administrative Region, China (CUHK 2300174 (Collaborative Research Fund, No. C5026-18GF)). We would like to thank our colleagues for their comments.

\bibliography{acl2020}
\bibliographystyle{acl_natbib}

\end{document}